\documentclass[letterpaper, 10 pt, conference]{ieeeconf}  

\IEEEoverridecommandlockouts                              

\usepackage{mathptmx}       
\usepackage{helvet}         
\usepackage{courier}        
\usepackage{type1cm}        
\usepackage{amsmath}
\usepackage{amssymb}
\usepackage{makeidx}         
\usepackage{graphicx}        
\usepackage{multicol}        
\usepackage[bottom]{footmisc}
\usepackage{enumerate}
\usepackage[all]{xy}
\usepackage{caption}
\usepackage{cite}
\usepackage[ruled]{algorithm2e}
\usepackage[usenames, dvipsnames]{color}
\usepackage{bm}
\usepackage{mathrsfs}
\usepackage{graphicx}
\usepackage{epsfig}
\usepackage{pdfpages}
\usepackage{wrapfig}
\usepackage{lipsum}
\usepackage{times}
\usepackage{setspace}
\usepackage{cancel}
\usepackage{wrapfig}
\usepackage[rightcaption]{sidecap}
\usepackage{adjustbox}
\usepackage{xspace}
\usepackage{float}
\usepackage{hyperref}

\usepackage{subfigure}

\newcommand{\ignore}[1]{}

\newcommand{\mbfdot}[1]{\dot{\mbf{#1}}}
\newcommand{\mbsdot}[1]{\dot{\mbs{#1}}}

\newcommand{\abs}[1]{\left\vert#1\right\vert} 

\newcommand{\bbm}{\begin{bmatrix}}
\newcommand{\ebm}{\end{bmatrix}}
\newcommand{\bma}[1]{\left[\begin{array}{#1}}
\newcommand{\ema}{\end{array}\right]}

\DeclareMathAlphabet{\mbf}{OT1}{ptm}{b}{n}
\newcommand{\mbs}[1]{{\boldsymbol{#1}}}

\newcommand{\mbfbar}[1]{{\bar{\mbf{#1}}}}

\newcommand{\beq}{\begin{equation}}
\newcommand{\eeq}{\end{equation}}
\newcommand{\bdis}{\begin{displaymath}}
\newcommand{\edis}{\end{displaymath}}
\newcommand{\beqarray}{\begin{eqnarray}}
\newcommand{\eeqarray}{\end{eqnarray}}
\newcommand{\beqarraynn}{\begin{eqnarray*}}
\newcommand{\eeqarraynn}{\end{eqnarray*}}

\newcommand{\p}{\partial}

\newcommand{\trans}{{\ensuremath{\mathsf{T}}}} 
\newboolean{draft-mode}
\setboolean{draft-mode}{true}
\newcommand{\sidenote}[1]{\ifthenelse{\boolean{draft-mode}}{\marginpar{\tiny\raggedright\textsf{\hspace{0pt}#1}}}{}}
\DeclareRobustCommand{\arnote}[1]{\ifthenelse{\boolean{draft-mode}}{\textcolor{blue}{\textbf{AR: #1}}}{}}
\DeclareRobustCommand{\ernote}[1]{\ifthenelse{\boolean{draft-mode}}{\textcolor{cyan}{\textbf{ER: #1}}}{}}
\DeclareRobustCommand{\fhnote}[1]{\ifthenelse{\boolean{draft-mode}}{\textcolor{red}{\textbf{FH: #1}}}{}}

\title{\LARGE \bf
Reactive Planar Manipulation with Convex Hybrid MPC
}

\author{\authorblockN{Francois Robert Hogan, Eudald Romo Grau, and Alberto Rodriguez}
  \authorblockA{Department of Mechanical Engineering ---
    Massachusetts Institute of Technology\\
    {\tt\small $<$fhogan,eudald,albertor$>$@mit.edu}}
    \thanks{This work was supported by
    NSF award [IIS-1637753] through the
    National Robotics Initiative.}}
    
\begin{document}

\maketitle
\thispagestyle{empty}
\pagestyle{empty}

\begin{abstract}

This paper presents a reactive controller for planar manipulation tasks that leverages machine learning to achieve real-time performance. The approach is based on a Model Predictive Control (MPC) formulation, where the goal is to find an optimal sequence of robot motions to achieve a desired object motion. Due to the multiple contact modes associated with frictional interactions, the resulting optimization program  suffers from combinatorial complexity when tasked with determining the optimal sequence of modes.

To overcome this difficulty, we formulate the search for the optimal mode sequences offline, separately from the search for optimal control inputs online. Using tools from machine learning, this leads to a convex hybrid MPC program that can be solved  in real-time. We validate our algorithm on a planar manipulation experimental setup where results show that the convex hybrid MPC formulation with learned modes achieves good closed-loop performance on a  trajectory tracking problem.

\end{abstract}

\section{INTRODUCTION}

While humans naturally make use of sensing and feedback when manipulating objects,  robot manipulators traditionally execute actions relying on open-loop control strategies.  Given the uncertainty associated with frictional contact interactions~\cite{Yu_IROS_2016,Bauza_ICRA_2017} and the inherent inaccuracies of contact models~\cite{Fazeli_ICRA_2017,Kolbert_ISER_2016}, 
the use of feedback can play an important role to address model uncertainty.
%
%
The long term goal of this work is to endow robots with real-time decision making capabilities to enable reactive manipulation.
 
\begin{figure}[t]
\centering
{
		\includegraphics[width=8cm]{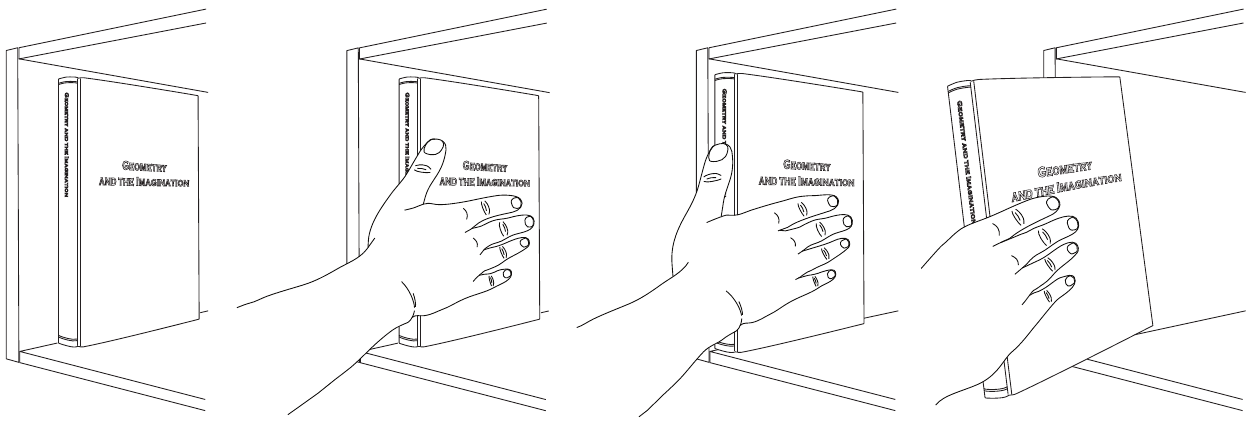} %
}
\centering
\caption{Animation of a hybrid 
manipulation task. The hand can interact with the book using one or many contact configurations and by exploiting different contact modalities, namely separation, sticking, and sliding. Figure adapted from \cite{hogan_WAFR_2016}.}
\label{fig:manipulation_book}
\end{figure}

This paper focuses on planar manipulation tasks where the physical interactions between manipulator, object, and environment can be modeled from first principles, using rigid body dynamics and Coulomb's frictional law. A major challenge concerning feedback controller design for systems involving contact interactions is the presence of hybridness and underactuation~\cite{hogan_WAFR_2016}. Hybridness refers to the fact that frictional interactions between manipulator and object exhibit different contact modes (e.g. contact/separation, sticking/sliding, etc), while underactuation is a result of the limited set of forces and torques that can be transmitted by the robot to the object through frictional interactions. Figure~\ref{fig:manipulation_book} shows an animation of a  hybrid  manipulation task that exploits multiple contact modalities.

This paper's main contribution is twofold. First, we present a controller design formulation that can be used to manipulate an object on a flat surface. The approach presented in this paper generalizes to multiple contact interactions between manipulator and object and for tracking  trajectories in the plane. 
Second, we introduce a method to determine an effective contact mode sequence that leads to a convex hybrid Model Predictive Control (MPC) formulation. Due to the hybridness associated with frictional interactions,  hybrid MPC formulations suffer from a combinatorial expansion due to unknown future contact interactions modes. To overcome this difficulty, we formulate the search for optimal modes separately from the search for optimal control inputs, by leveraging machine learning methods to select mode sequences from prior experience.  Once the mode sequences are selected, the control problem reduces to solving a convex quadratic program, which can be achieved at very high frequency.

\section{RELATED WORK}

The mechanics of planar pushing manipulation tasks were first described by~\cite{Mason_1986}. \cite{Goyal_1991}\, introduced the concept of the \emph{limit surface}, a useful geometric representation which maps the applied frictional forces on an object to its instantaneous velocity. Under the assumption of quasi-static interactions, the limit surface has been successfully used in simulation~\cite{Mason_1986}, planning~\cite{Lynch_1996}, state estimation~\cite{Yu_IROS_2015}, and feedback control~\cite{Lynch_1992, hogan_WAFR_2016, Woodruff_ICRA_2017} applications. Due to the high computational costs associated with building a true limit surface, \cite{Cutkosky_1991} proposed an ellipsoidal approximation which  yields invertible models from force to motion. Recently, \cite{Zhou_ICRA_2016} exploited the convex properties of the limit surface to develop an efficient data-driven algorithm for its construction from contact interactions. 
 
There is an ongoing effort to find planning frameworks that can effectively handle  the underactuation and hybridness associated with contact models. \cite{Chavan_ISRR_2017} used sampled-based planning algorithms to plan robot motions for in-hand manipulation tasks. \cite{Posa_2014} developed a nonlinear trajectory optimization framework that includes frictional forces as decision variables within a nonlinear program and makes use of complementary constraints to encode the different contact interaction modes of the system. Another approach that shows  promise is based on Differential Dynamic Programming (DDP), which iteratively approximates locally-quadratic models of the dynamics and cost functions to find a locally optimal path. Most approaches in this line rely on approximating discontinuous dynamics with continuous relaxations. \cite{Tassa_2012} used penalty methods to smooth contact models while \cite{Pajarinen_2017} modeled the discontinuous dynamics of the system with  mixture models.

The application of model-based feedback control to contact rich tasks has been limited to a small number of  applications \cite{Lynch_2012, Posa_ICRA_2016,Woodruff_ICRA_2017}. The control strategies presented in the aforementioned papers are applied to systems with an a priori knowledge of the contact mode sequencing. In \cite{hogan_WAFR_2016}, a feedback controller design is presented for the pusher-slider system using a Model Predictive Control framework, where a set of contact mode
schedules are chosen  such that they span a  number of dynamic behaviors likely to occur. This method has been shown to work experimentally but requires heuristic methods in order to design  candidate contact mode schedules. This paper aims to eliminate the need for contact mode enumeration based on human intuition by developing an algorithm that systematically selects  contact modes sequences. 

\section{NOMENCLATURE}

The notation used in the paper is described below:
\begin{itemize}   	
	\item $H$: Convex set representing the limit surface.
	\item $\mbf{w} = [f_x\,\,\,\,f_y\,\,\,\,\tau]^\trans$: Applied wrench on the object resolved in the body frame.
	\item $\mbf{t} = [v_x\,\,\,\,v_y\,\,\,\,\omega]^\trans$: Object twist  resolved in the body frame.
	\item $\mbf{J}_c$: Jacobian matrix associated with the contact point $c$  resolved in the body frame.
	\item $\mbf{N} = [\mbf{n}_1^\trans\,\,\,\,\hdots\,\,\,\,\mbf{n}_C^\trans]^\trans$: Matrix of object normal vectors at contact points resolved in body frame.
	\item  $\mbf{D} = [\mbf{d}_1^\trans\,\,\,\,\hdots\,\,\,\,\mbf{d}_C^\trans]^\trans$: Matrix of object tangent vectors at contact points resolved in body frame.
	\item $\mbf{f}_n = [{f}_{n,1}\,\,\,\,\hdots\,\,\,\,{f}_{n,C}]^\trans$: Vector of applied normal forces at contact points resolved in body frame.
	\item $\mbf{f}_t = [{f}_{t,1}\,\,\,\,\hdots\,\,\,\,{f}_{t,C}]^\trans$: Vector of applied tangential forces at contact points resolved in body frame.
	\item $\mbs{\phi} = [{\phi}_{1}\,\,\,\,\hdots\,\,\,\,{\phi}_{C}]^\trans$: Vector of relative angles of pusher relative to body frame.
	\item $\mbf{x} = [x\,\,\,\,y\,\,\,\,\theta\,\,\,\,\mbs{\phi}^\trans]^\trans$: System state vector.
	\item $\mbf{u}_c = [
	{f}_{n,c}\,\,\,\,{f}_{t,c}\,\,\,\,\dot{\phi}_c
	]^\trans$: Vector of control inputs at contact point $c$.
	\item $\mbf{u}_f = [
	\mbf{f}_{n}^\trans\,\,\,\,\mbf{f}_{t}^\trans
	]^\trans$: Vector of commanded reaction forces.
    \item $\mbf{u}_\phi =
	\mbsdot{\phi}$: Vector of commanded angular velocities.
	\item $\mbf{u} = [
	\mbf{u}_{f}^\trans\,\,\,\,\mbf{u}_{\phi}^\trans
	]^\trans$: Control input.
\end{itemize}

\section{Planar Pushing Model}
\label{sec:planer_model}
This section introduces the motion model for planar manipulation tasks that is general to an arbitrary number of contact points and arbitrary object shapes.
%
%
We adapt the modeling of planar pushing interactions from \cite{Zhou_RSS_2017} that we briefly summarize below. All contacts in this work assume Coulomb friction interactions~\cite{book_Mason}, uniform pressure distribution, and quasi-static interactions, where the inertial forces of the object are negligible. 

\subsection{Motion Model}
\label{sec:motion_model}

The limit surface is a geometric representation that describes the relationship between the applied force on an object and its instantaneous velocity. In this paper, we use the ellipsoidal approximation to the limit surface \cite{Cutkosky_1991, Zhou_RSS_2017} due its simplicity and invertibility properties. The ellipsoidal limit surface can be expressed in convex quadratic form as
$ H(\mbf{w})=\frac{1}{2}\mbf{w}^\trans\mbf{A}\mbf{w}$. 
By the principle of maximal dissipation~\cite{Goyal_1991}, the object twist is perpendicular to the limit surface for a given wrench
\beqarray
\mbf{t} &=&  \mbs{\nabla}H(\mbf{w}) =  \mbf{A} \mbf{w},
\eeqarray
where the applied frictional wrench is 
\beq
\mbf{w}=\sum_{c=1}^C \mbf{J}_{c}^\trans\left(
\mbf{n}_c  f_{n,c}+ \mbf{d}_{c}{f}_{t,c}
\right).
\eeq

\begin{figure}[h]
\centering
{
		\includegraphics[width=3.3cm]{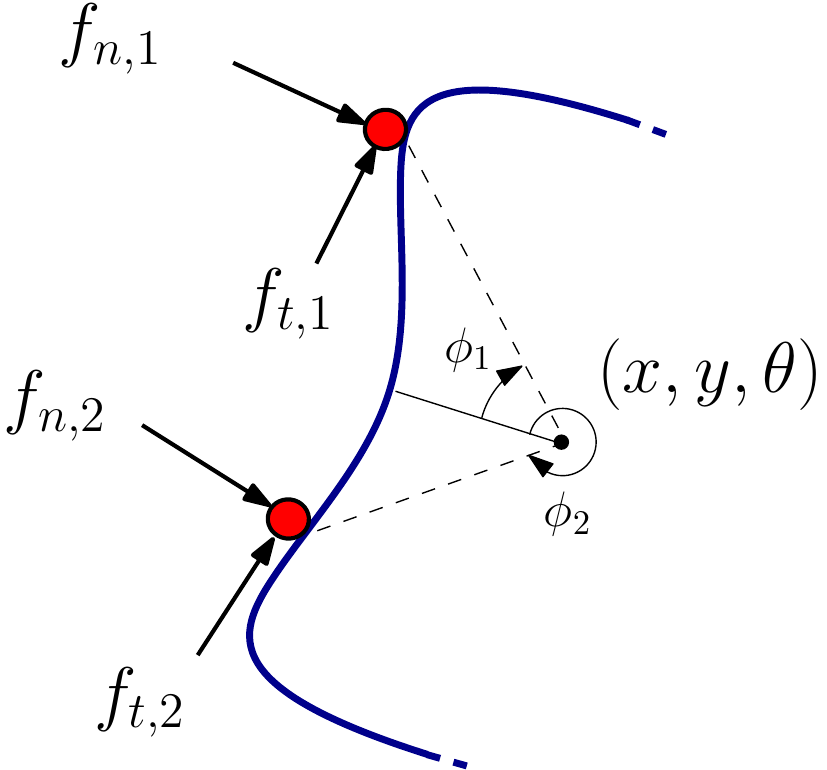} %
}
\centering
\caption{Free body diagram of a sliding object with $c=2$ contact points.} \label{fig:2pt_general_shape}
\end{figure}

Consider the planar manipulation task with multiple contact points shown in Fig.~\ref{fig:2pt_general_shape}. The unconstrained motion equations of the system can be expressed as
\beq
\mbfdot{x} = \mbf{f}(\mbf{x},\mbf{u}) =  \bma{cc}
\mbf{R}\,\mbf{t} \\ 
\mbf{u_\phi}
\ema,
\hspace{3mm}
\mbf{R} = \bma{ccc}
\cos\theta&-\sin\theta&0\\
\sin\theta&\cos\theta&0\\
0&0&1
\ema,
\label{eq:motion_equations}
\eeq
 assuming that all points maintain contact with the sliding object.

\subsection{Frictional Constraints}
\label{sec:contact_constraints}

The motion equations in \eqref{eq:motion_equations} do not enforce that the reaction forces between manipulator and sliding object are feasible. To ensure that the motion equations are associated with physically reasonable behavior, we must impose constraints on the control input   $\mbf{u}$, ensuring  that the motion model obeys contact interactions laws. Due to the hybrid nature of contact, the physical constraints that dictate the magnitude and direction of the frictional forces vary with the contact interaction mode. In accordance with Coulomb's frictional law, the following constraints on the inputs must always be satisfied independently of the contact mode:
\begin{figure}[h]
\begin{minipage}[b]{0.55\linewidth}
\hspace{-20mm}
\centering
\includegraphics[width=1.5cm]{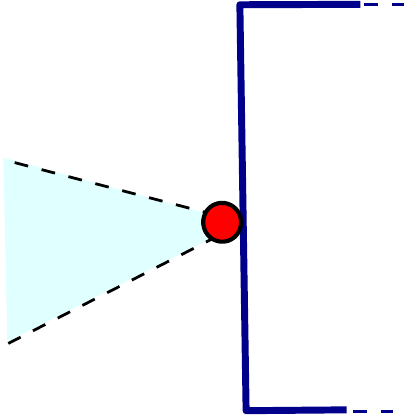}
\caption{Friction cone constraint. The applied force must remain within the blue shaded region.} \label{fig:mode_ind}
\end{minipage}
\hspace{0.5cm}
\begin{minipage}[b]{0.2\linewidth}
\beqarray
\hspace{-10mm}
\mathcal{C}_0: 
\begin{cases}
f_{n,c}\geq0, \label{uni_1}\\ 
\abs{f_{t,c}}\geq\mu_p f_{n,c}\label{uni_3}, 
\end{cases}
\eeqarray
\vspace{17mm}
\end{minipage}
\end{figure}
\vspace{-2mm}

\noindent implying the pusher can only exert a compressive force on the object and that the net frictional force applied on the object remains within the bounds of the friction cone  in Fig.~\ref{fig:mode_ind}. 
In addition, we must enforce constraints that depend on the contact interaction mode.

\begin{figure}[h]
\centering
\subfigure[Sticking. The relative velocity between the pusher and object is zero.]
{
			\includegraphics[width=2cm]{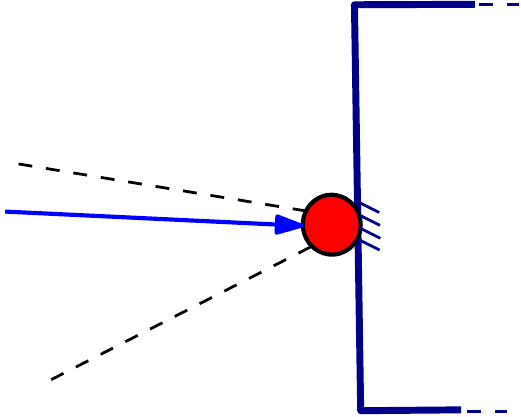} 
		\label{fig:sticking}
}
\hspace{5mm}
\subfigure[Sliding left. The frictional force lies on the lower boundary of the friction cone.]
{
			\includegraphics[width=2cm]{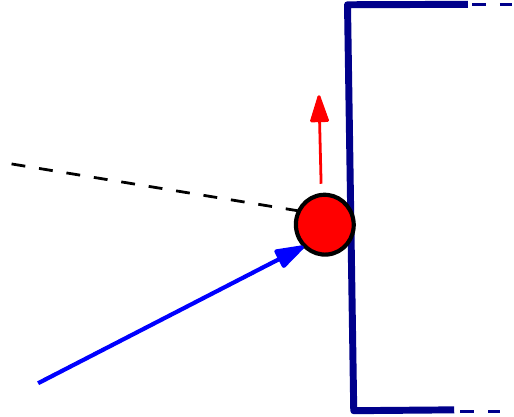} 
		\label{fig:sliding_left}
}
\hspace{5mm}
\subfigure[Sliding right. The frictional force lies on the upper boundary of the friction cone.]
{
			\includegraphics[width=2cm]{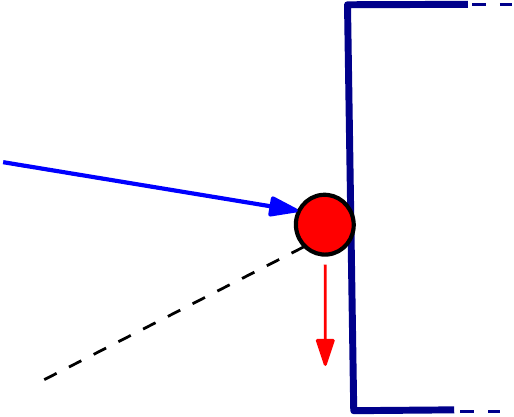} 
		\label{fig:sliding_right}
}
\centering
\caption{Mode dependent constraints following Coulomb's frictional interaction law.}	
\label{model}
\end{figure}

\textbf{Sticking} When the pusher is sticking relative to the object, the tangential velocity is stationary, as in Fig.~\ref{fig:sticking}

\beq
\mathcal{C}_{1}: \hspace{2mm}\dot{\phi}_c = 0.
\eeq

\textbf{Sliding Left} When the pusher is sliding left relative to the object, the tangential velocity is strictly positive and the frictional force must remain on the right hand side of the friction cone, as in Fig.~\ref{fig:sliding_left}
\beqarray
\mathcal{C}_2: 
\begin{cases}
\hspace{2mm}
\dot{\phi}_c > 0, \\ 
f_{t,c} = \mu_p f_{n,c}.
\end{cases}
\eeqarray

\textbf{Sliding Right} When the pusher is sliding right relative to the object, the tangential velocity is strictly negative and the frictional force is constrained to remain on the left hand side of the friction cone, as in Fig.~\ref{fig:sliding_right}

\beqarray
\mathcal{C}_3: 
\begin{cases}
\dot{\phi}_c < 0, \\ 
f_{t,c} = -\mu_p f_{n,c}.
\end{cases}
\eeqarray

\section{HYBRID MODEL-PREDICTIVE CONTROL}
\label{sec:controller_design}
This section presents a controller design framework for planar manipulation tasks. The proposed controller aims to stabilize the motion of a given object about a nominal trajectory. The approach presented follows a Model Predictive Control (MPC) formulation where the goal is to determine a sequence of control inputs over a receding horizon to minimize the error between the manipulated object and its desired motion. Due to the hybridness associated with Coulomb's frictional law, the optimization program takes the form of a mixed-integer quadratic program (MIQP).

\vspace{2mm}
\noindent \textbf{Optimization Problem MPC (MIQP)}: Given current error state $\mbfbar{x}_0$ and nominal trajectory ($\mbf{x}_i^\star$, $\mbf{u}_i^\star$), solve 
\vspace{-2mm}
\beq
\begin{aligned}
& \underset{\mbfbar{x}_i,\hspace{1mm} \mbfbar{u}_i,\hspace{1mm}  \mbf{z}_{i}}{\text{min}}
& &  \mbfbar{x}_N^\trans\mbf{Q}_N\mbfbar{x}_N + \sum_{i=0}^{N-1} \left(\mbfbar{x}_{i+1}^\trans\mbf{Q}\mbfbar{x}_{i+1} + \mbfbar{u}_i^\trans \mbf{R}\mbfbar{u}_i
+ \mbf{z}_i^\trans \mbf{W}\mbf{z}_i
\right) \\
& \text{subject to}
& & {\mbfbar{x}}_{i+1} ={\mbfbar{x}}_{i}+ h\left[\mbf{A}_i{\mbfbar{x}}_{i}+ \mbf{B}_i{\mbfbar{u}}_{i}\right], \\
&&& \mbfbar{u}_{c,i} \in \mathcal{C}_0, 
 \\
&&& 
\mbfbar{u}_{c,i} \in \mathcal{C}_1 \hspace{5mm} \text{if  $c$ is sticking } (\text{i.e., } z_{1c,i}=1),\\
&&& 
\mbfbar{u}_{c,i} \in \mathcal{C}_2 \hspace{5mm} \text{if  $c$ is sliding left } (\text{i.e., }  z_{2c,i}=1),  \\ 
&&& 
\mbfbar{u}_{c,i} \in \mathcal{C}_3 \hspace{5mm} \text{if  $c$ is sliding right } (\text{i.e., }  z_{3c,i}=1),\\
&&& z_{1c,i} + z_{2c,i} + z_{3c,i}  =  1,
\end{aligned}
\label{mpc_miqp}
\eeq
\vspace{-1mm}

\noindent with $\mbfbar{x}_i = \mbf{x}_i - \mbf{x}_i^{\star}$, $\mbfbar{u}_i = \mbf{u}_i - \mbf{u}_i^{\star}$, and $\mbf{z}_{i} = [z_{1c,i}, z_{2c,i}, z_{3c,i}]^\trans$.   The terms $\mbf{Q}$, $\mbf{Q}_N$,  $\mbf{R}$, and $\mbf{W}$ denote weight matrices associated with the error state, final error state,  control input, and contact modes, respectively. We constrain the search to the linearized dynamics of the system and the contact constraints presented in sections~\ref{sec:planer_model}, where $\mbf{A}_i  = \frac{\p \mbf{f}}{\p \mbf{x}}|_{\mbf{x}_i^\star, \mbf{u}_i^{\star}}$ and $\mbf{B}_i  = \frac{\p \mbf{f}}{\p \mbf{u}}|_{\mbf{x}_i^\star, \mbf{u}_i^{\star}}$.

\begin{figure}[h]
\centering
{
  \includegraphics[width=8.5cm]{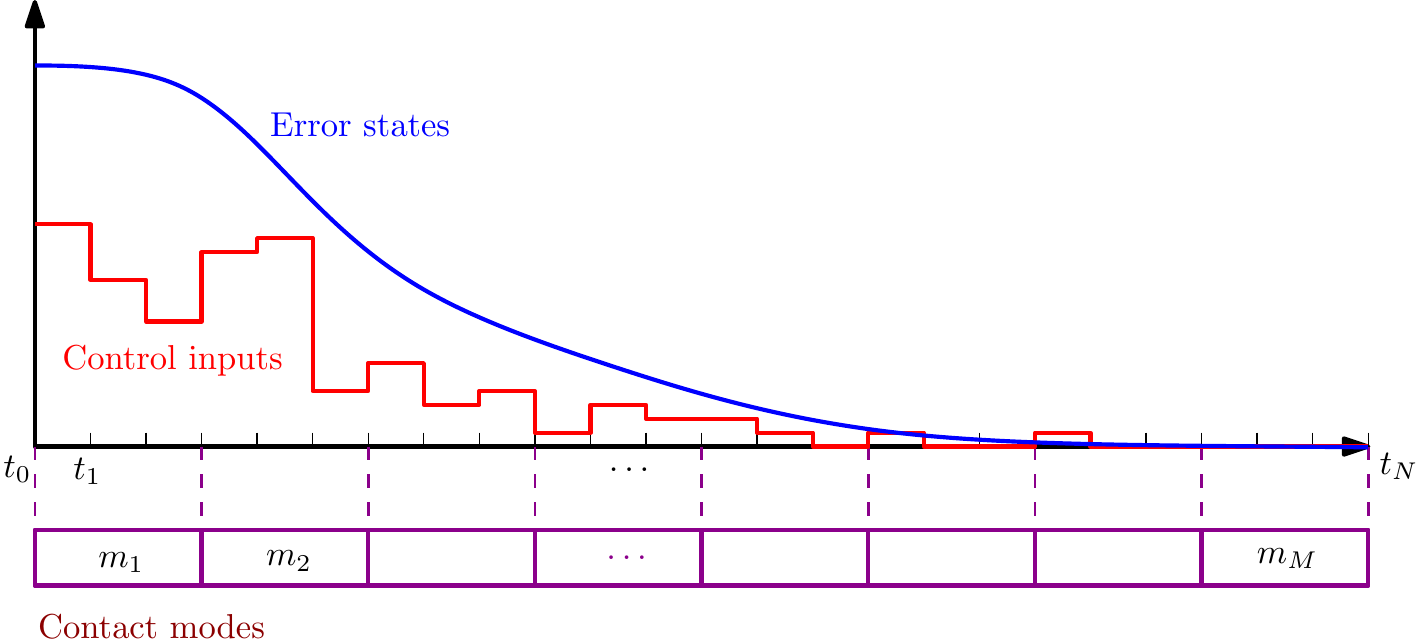} %
}
\centering
\caption{Hybrid MPC framework.  A sequence of  control inputs is computed that will drive the predicted states to the reference trajectory while simultaneously finding the schedule of optimal hybrid mode transitions $\mbf{m} = \{m_1, \hdots, m_M\}$. The control input $\mbfbar{u}_0 + \mbf{u}_0^\star$ is applied to the system.} \label{fig:mode_division}
\end{figure}

We introduce integer variables into the optimization program to denote the hybrid mode that is active at time step $i$. The integer variables $
z_{1c,i}$,
$z_{2c,i}$, and 
$z_{3c,i}$ $\in \{0,1 \},
$
denote sticking, sliding left, and sliding right for contact point $c$,  respectively, where the integer variable takes the value of $1$ if a contact mode is active and $0$ otherwise. 
We enforce that the sum of integers values must be unity at each time step to ensure that only one mode can be active at a time.  

To speed up computation, it is often practical to constraint adjacent time steps within a prediction horizon to have the same contact mode. This is shown in Fig.~\ref{fig:mode_division}, where the agglomerated mode sequence $\mbf{m} = \{m_1, \hdots, m_M\}$ is introduced, with  $m_m \in \{ 1,2,3 \}$ denoting sticking, sliding left, and sliding right.

\section{Offline Mode Schedule Learning}

\begin{figure}[h]
\centering
{
  \includegraphics[width=6.5cm]{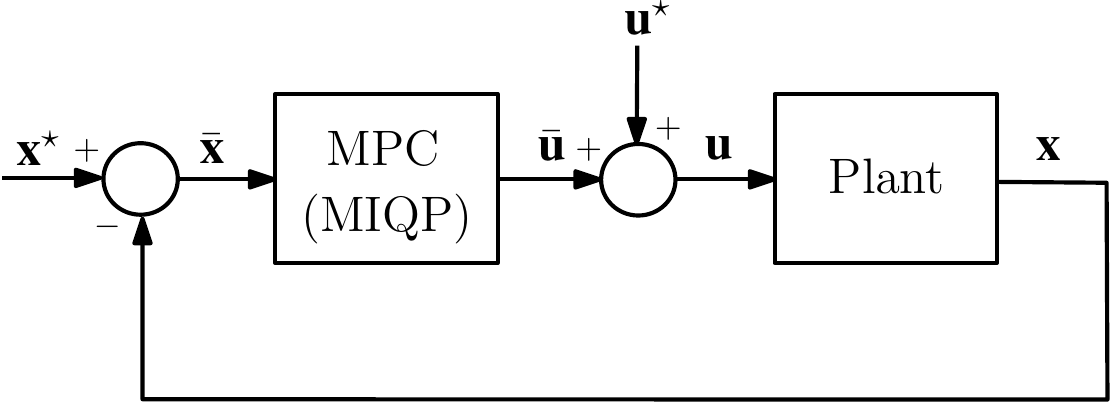} %
}
\centering
\caption{Block diagram of hybrid  controller design described in Eq.~\eqref{mpc_miqp}. The resulting MPC controller design is a non-convex mixed-integer quadratic program.} \label{fig:hybrid_mpc_block}
\end{figure}

We can visualize the feedback control architecture proposed in Section~\ref{sec:controller_design} in block diagram form in Fig.~\ref{fig:hybrid_mpc_block}. Due to the non-convex nature of integer variables in Eq.~\eqref{mpc_miqp}, the solution time is typically slow and not appropriate for high bandwidth feedback control applications.
In an effort to increase the  control bandwidth, we aim to offload as much of the computational costs offline as possible. To accomplish this, we present a novel formulation that separates the search for the mode schedule selection from the optimal control sequence.
\vspace{0mm}

\begin{figure}[h]
\centering
{
  \includegraphics[width=7.5cm]{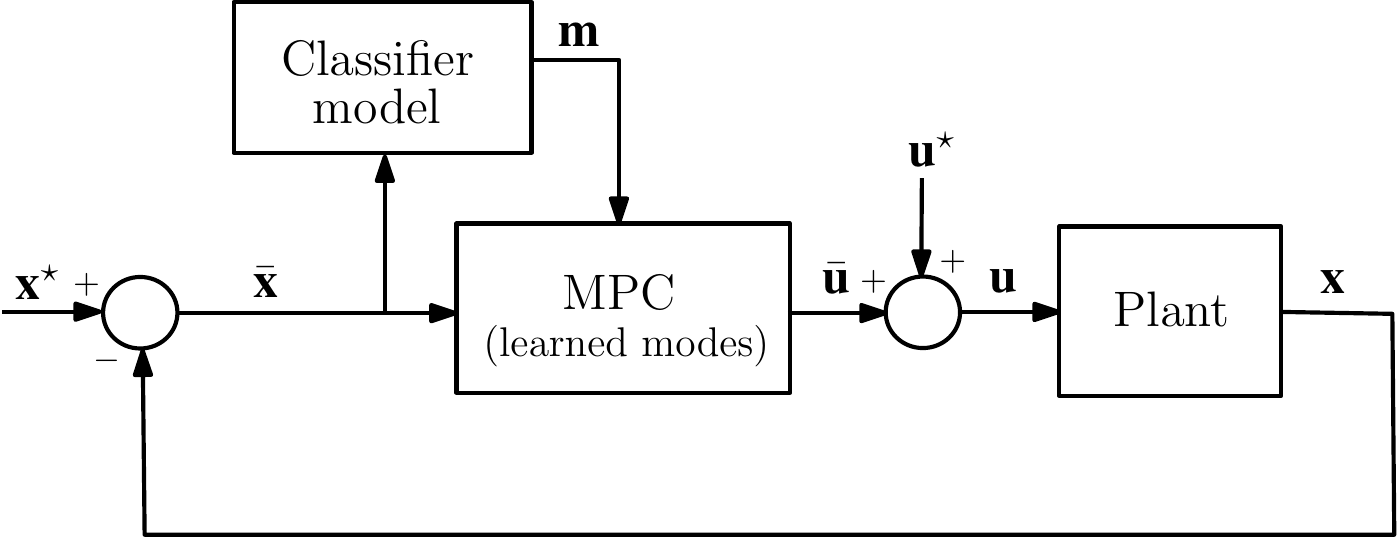} %
}
\centering
\caption{Block diagram of hybrid  controller design with learned mode schedule classifier. The resulting MPC controller design is a convex quadratic program.} \label{fig:learned_mpc_block}
\end{figure}

\begin{figure}[h]
\centering
{
  \includegraphics[width=8.6cm]{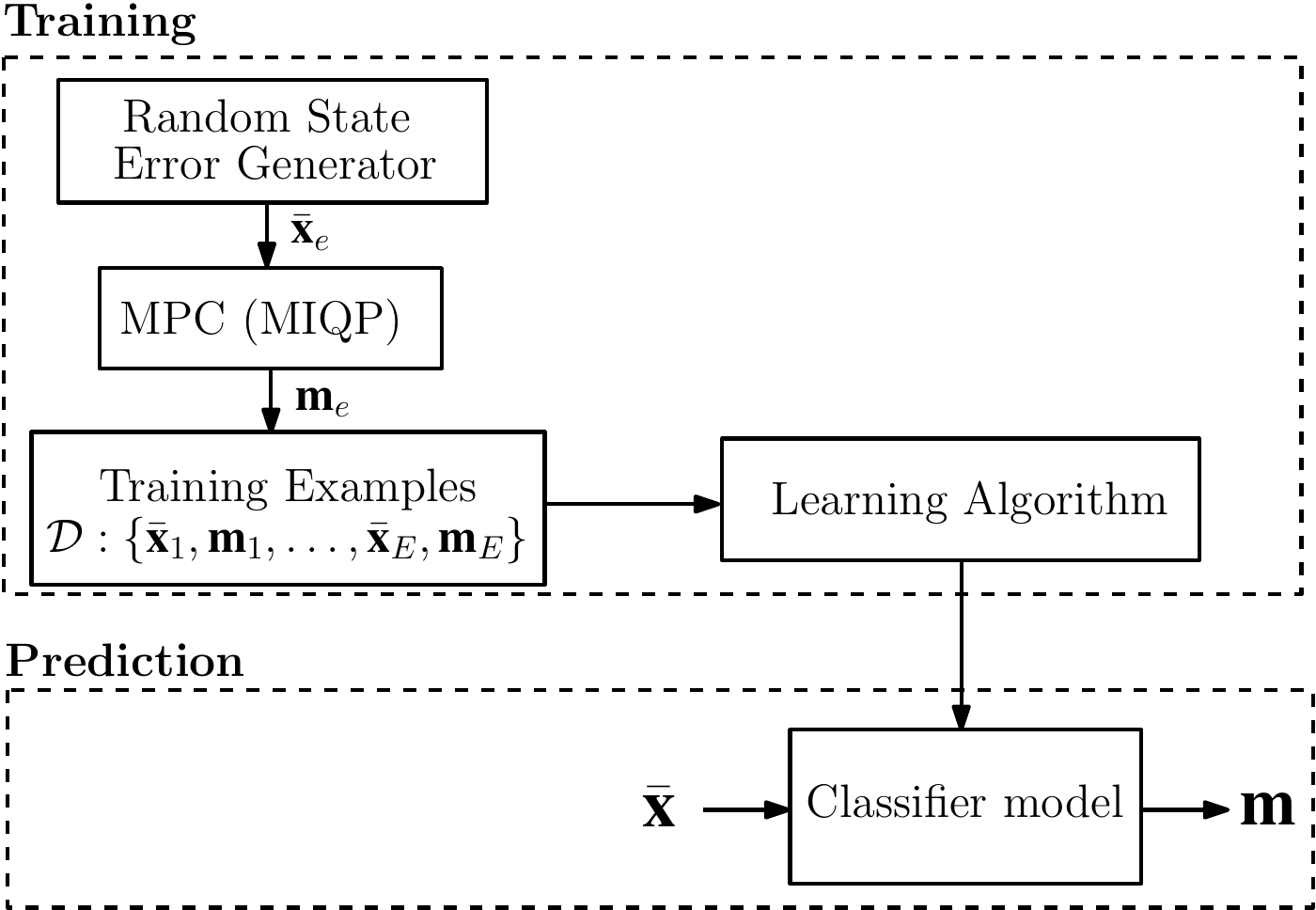} %
}
\centering
\caption{Supervised learning framework for mode schedule selection. A dataset of $E$ labelled datapoints is generated using the the MPC (MIQP) formulation. From the training examples, a classifier is trained to return the mode schedule based on the state error vector.} \label{fig:learning_framework}
\end{figure}

Consider the controller design architecture proposed in Figure~\ref{fig:learned_mpc_block}. Suppose that given the state error $\mbfbar{x}$, we had access to an oracle function that returned an effective mode schedule $\mbf{m} = \{m_1, m_2, \hdots, m_N\}$ to be enforced during the prediction horizon. 
\vspace{0mm}
%
Although we do not have direct access to a real-time function that determines the optimal mode schedule, we can query the mixed-integer MPC program as much as desired offline to find optimal mode sequences given error state inputs. This formulation lends itself well to a supervised learning setting, where the objective is to train a  classifier model that can select an effective mode schedule given the error state. We present the learning framework used to design the classifier model shown  in Fig.~\ref{fig:learning_framework}. Using the hybrid MPC (MIQP) formulation presented in Eq.~\eqref{mpc_miqp}, we generate a dataset of $E$ training example $\{\mbfbar{x}_e, \mbf{m}_e \}$, where $\mbf{m}_e$ represents the mode schedule associated with the $e^{th}$ datapoint. The purpose of the machine learning algorithm is to train a candidate classifier model that minimizes the cross-entropy error function of the labelled training set.  This new hybrid control architecture leads to a convex optimization program with a prescribed mode sequence and is referred to as MPC (learned modes). 

\vspace{2mm}
\noindent \textbf{Optimization Problem MPC (learned modes)}: Given current error state $\mbfbar{x}_0$, nominal trajectory ($\mbf{x}_i^\star$, $\mbf{u}_i^\star$), and mode schedule $\mbf{m}$, solve 
\vspace{-2mm}
\bdis
\begin{aligned}
& \underset{\mbfbar{x}_i,\hspace{1mm} \mbfbar{u}_i}
{\text{min}}
& &  \mbfbar{x}_N^\trans\mbf{Q}_N\mbfbar{x}_N + \sum_{i=0}^{N-1} \left(\mbfbar{x}_{i+1}^\trans\mbf{Q}\mbfbar{x}_{i+1} + \mbfbar{u}_i^\trans \mbf{R}\mbfbar{u}_i
\right) \\
& \text{subject to}
& & {\mbfbar{x}}_{i+1} ={\mbfbar{x}}_{i}+ h\left[\mbf{A}_i{\mbfbar{x}}_{i}+ \mbf{B}_i{\mbfbar{u}}_{i}\right], \\
&&& \mbfbar{u}_{c,i} \in \mathcal{C}_0, 
 \\
&&& 
\mbfbar{u}_{c,i} \in \mathcal{C}_1 \hspace{5mm} \text{if  $c$ is sticking},\\
&&& 
\mbfbar{u}_{c,i} \in \mathcal{C}_2 \hspace{5mm} \text{if  $c$ is sliding left},  \\ 
&&& 
\mbfbar{u}_{c,i} \in \mathcal{C}_3 \hspace{5mm} \text{if  $c$ is sliding right}.
\end{aligned}
\edis

The main attraction of this approach is to convert a non-convex mixed-integer quadratic program into a convex quadratic program that can be solved in real-time. 
%
\vspace{0mm}

\section{Results}
\label{sec:results}

In this section, we implement  the controller design presented in Section~\ref{sec:controller_design} along with the planar pushing model described in Section~\ref{sec:motion_model} on a planar pushing experimental setup. Videos of the experiments can be found at \url{https://youtu.be/bMMlkyue_ZU}. In both of the experiments considered in this section, we parametrize the classifier model introduced  in Fig.~\ref{fig:learned_mpc_block} with a  neural network   with the properties summarized in Table 1.
\begin{table}[t]
\caption{Experiment parameters.}
\vspace{-5mm}
\begin{center}
\label{table:table1}
\begin{tabular}{l l l l l}
& &  \\ 
 Property & Symbol  & Value\\
\hline
Coefficient of friction (pusher-slider) & $\mu_p$ &  $ 0.3 $ \\
Coefficient of friction (slider-table) & $\mu_g$ &  $ 0.35 $ \\
Mass of slider (experiment A), $kg$ & m & 0.827 \\
Object radius (experiment A), $m$ & r & 0.045 \\
Mass of slider (experiment B), $kg$ & m & 0.827 \\
Object side length (experiment B), $m$ & a & 0.09 \\
Line pusher width (experiment B), $m$ & d & 0.03
\end{tabular}
\end{center}
\end{table}

\subsection{Case Study A:  Single Point Pushing}
\label{sec:point_pushing}

\begin{figure}[h]
\centering
{
  \includegraphics[width=8.5cm]{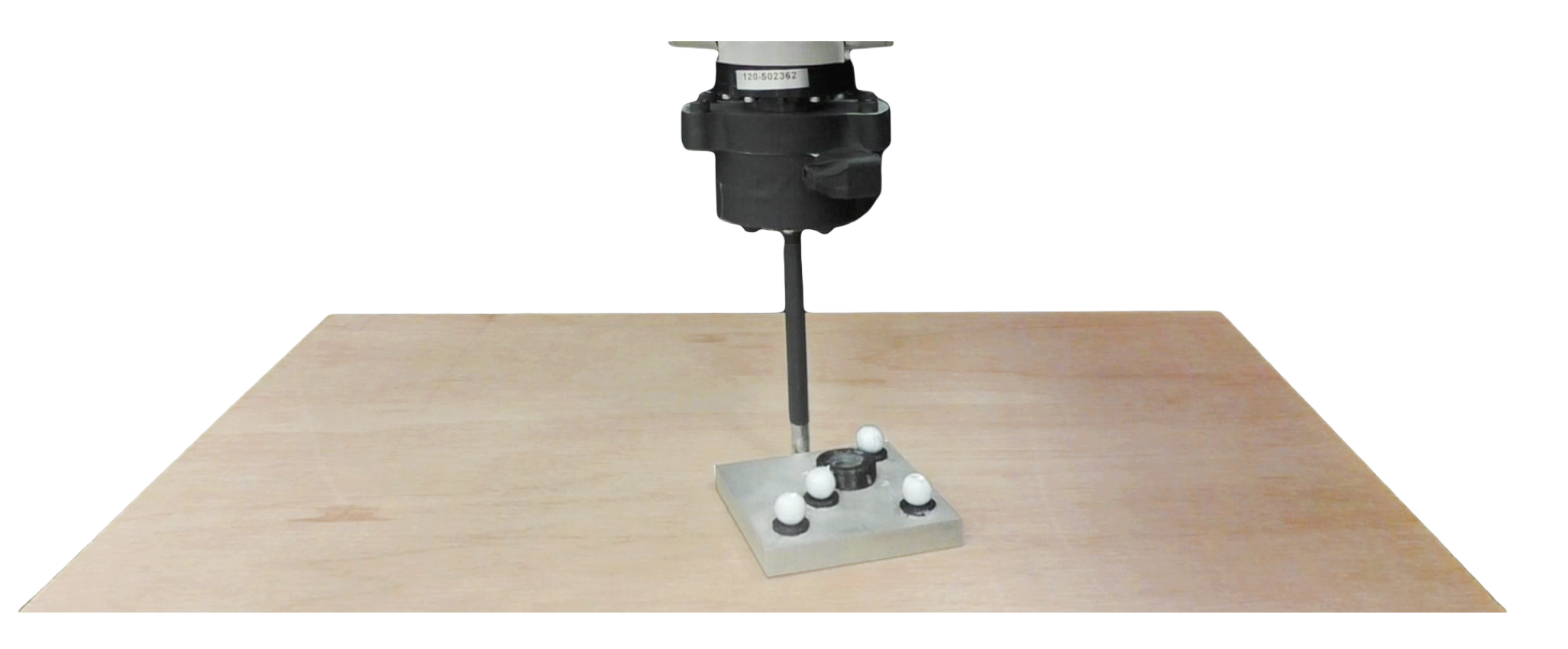} %
}
\centering
\caption{Experimental setup for point pusher. }
\label{fig:exp_setup}
\end{figure}

First, we investigate the performance of the controller design in Fig.~\ref{fig:learned_mpc_block} on a planar manipulation system where the goal is to track a 2d trajectory in the shape of an $8$ shaped trajectory defined by two circles of  radii $0.15$ meters at a constant velocity of $v=0.05$ [m/s]. We build the classifier model following the learning framework displayed in Fig.~\ref{fig:learning_framework}, where the training examples are generated using the  MPC (MIQP) program presented in  Eq.~\eqref{mpc_miqp}. 
\begin{figure}[h]
\centering
{
  \includegraphics[width=6cm]{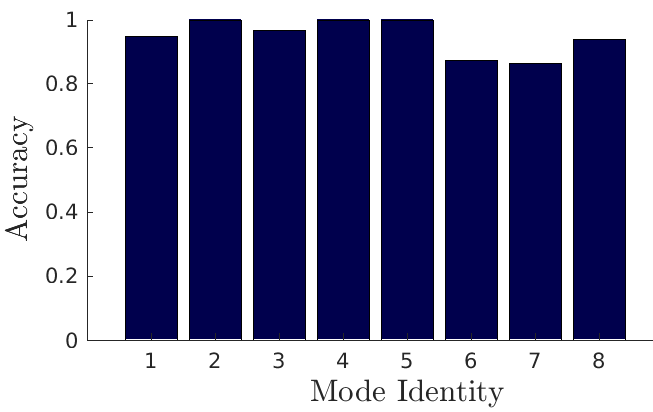} %
}
\centering
\caption{Accuracy results of the neural network predictions on a validation set of $58815$ labelled data points. We evaluate the performance on each mode separately, as defined in Fig.~\ref{fig:mode_division}. }
\label{fig:circular_trajectory}
\end{figure}
We parametrize the classifier using a  neural network, as detailed in Table 1. The physical properties of the circular object along with the frictional properties of the system are in Table 2. The controller design parameters used in the numerical simulations are   $h=0.3$ seconds, $N=35$, $\mbf{Q} = 10~\text{diag}\{3,3,0.1,0\}$, $\mbf{Q}_N = 2000~\text{diag}\{3,3,0.1,0\}$, and $\mbf{R} =0.5~\text{diag}\{1,1,0.01\}$. The prediction horizon is split into $8$ parts during which the contact modes are held constant. The number of time steps  associated with each contact mode section $m_m$ is $\{1,5,5,5,5,5,5,4\}$  with the associated contact mode weight matrix $\mbf{W} = 0.1~\text{diag}\{0,3,1,1,1,0,0,0\}$.

\begin{table}[t]
\caption{Neural network parameters.}
\vspace{-5mm}
\begin{center}
\label{table:table1}
\begin{tabular}{l l l l}
& &  \\ 
 Property   & Value\\
\hline
Number of  hidden layers      & 3       \\
Neurons in hidden layer 1       & 32         \\
Neurons in hidden layer 2       & 50         \\
Neurons in hidden layer 3       & 50         \\
Activation functions       & ReLu         \\
Output layer       & Softmax         \\
Loss function & Cross entropy           \\
\end{tabular}
\end{center}
\end{table}
\begin{figure}[h]
\centering
\hspace{-5mm}
\subfigure[Optimal contact mode (MIQP).]
{
\includegraphics[width=3.83cm]{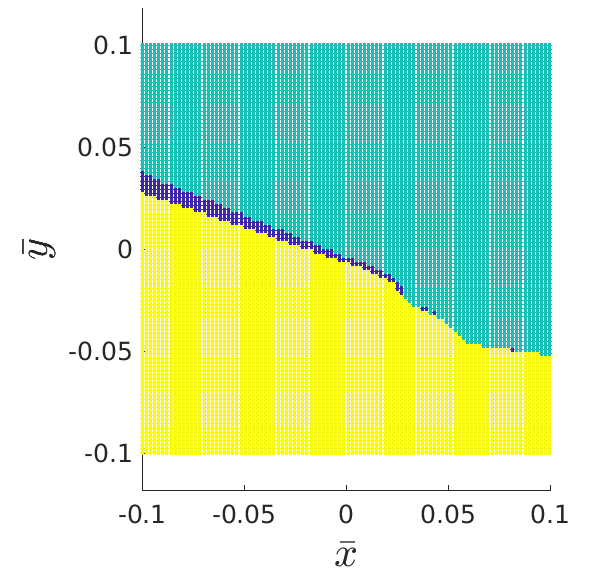}
		\label{fig:center}
}
\hspace{-4mm}
\subfigure[Classifier prediction.]
{
			\includegraphics[width=4.77cm]{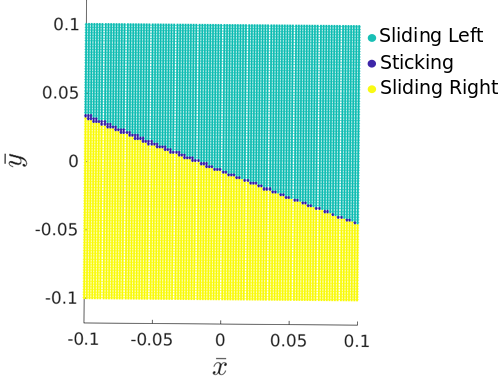} 
		\label{fig:center_learned}
}
\centering
\caption{Optimal contact mode for the first time step as a function of initial state errors in $x$ and $y$ while holding $\theta = 5$ [deg] and $\phi = 0$ [deg]. The classifier model captures the important trends of the MPC (MIQP) optimal mode solutions.}	
\label{fig:clusters}
\end{figure}

Figure~\ref{fig:circular_trajectory} shows the prediction accuracy of the  neural network trained on $117630$ labelled data points on a validation set of $58815$ labelled data points both generated using sampling the error state from a normal distribution with standard deviation $[0.03\,\, 0.03\,\, .4\,\, 0.025]$. We evaluate the performance on each mode individually, as defined in Fig.~\ref{fig:mode_division}. 

Figure~\ref{fig:clusters} compares the optimal contact mode associated with  the first time step of the MPC (MIQP) with the prediction made by the classifier. The contact modes are generated  as a function of initial state errors in $x$ and $y$ while holding $\theta$ and $\phi$ constant at $5$ degrees. The regions shown in green, yellow, and blue denote the regions where the options actions are sliding left, sliding right, or sticking. From the figure, we notice that the optimal contact modes are separated into distinct region, thus justifying the search for contact mode  as a classification problem and facilitating the learning process.  We notice that the  classifier model succeed in capturing the important trends of the MPC (MIQP)  solutions.

\begin{figure*}[t]%
\centering
\subfigure[Tracking of the $8$ track  for  $7$ consecutive laps. The black line represents the desired trajectory and the blue lines track the center of mass of the object.]{%
\label{fig:point_pusher_experiments_first}%
\includegraphics[width=8cm]{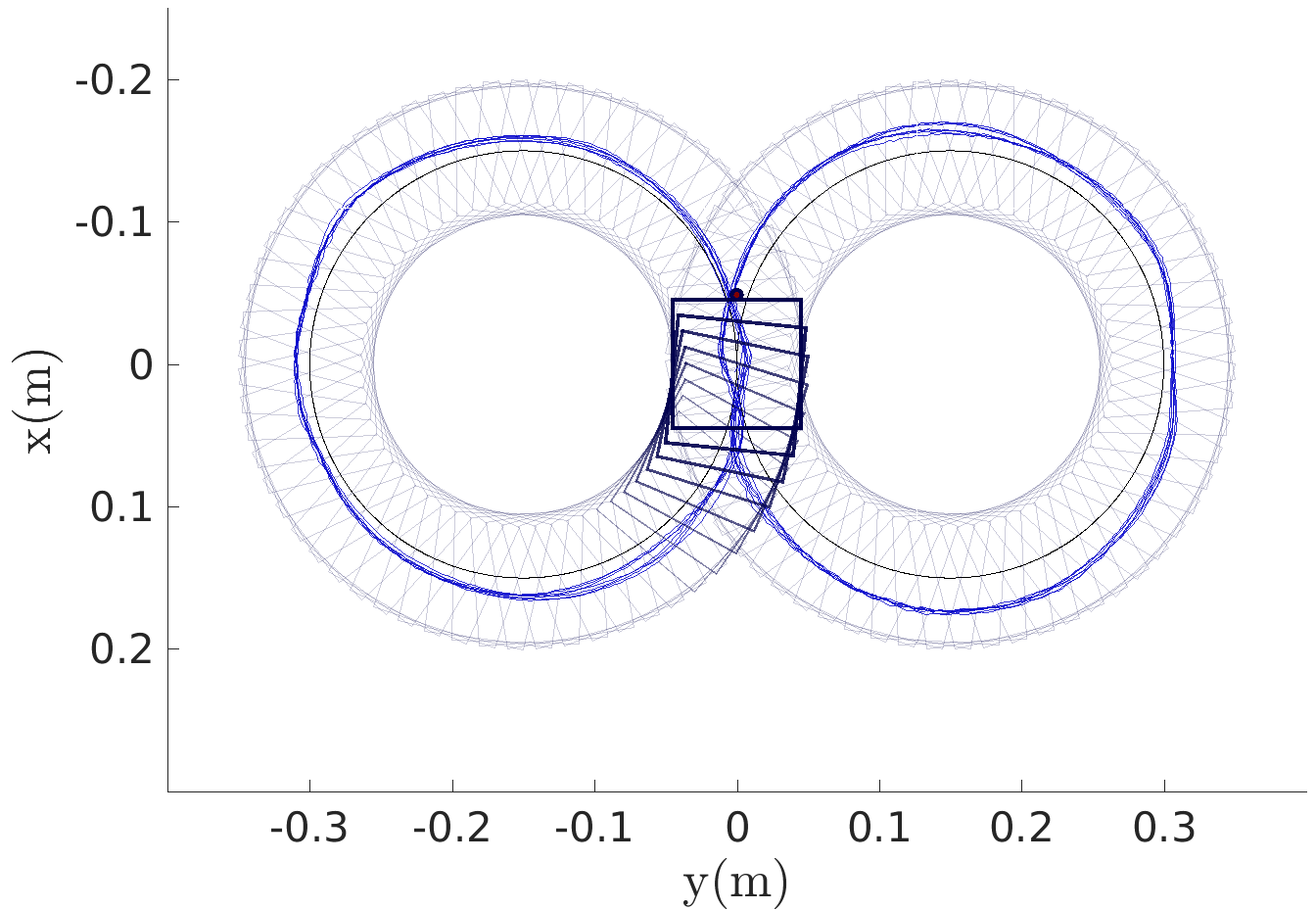}}%
\qquad
\subfigure[Tracking of the $8$ track with  external perturbations for a single laps. The black line represents the desired trajectory and the hand represents the locations and directions in which  the perturbations were applied.]{%
\label{fig:point_pusher_experiments_second}%
\includegraphics[width=8cm]{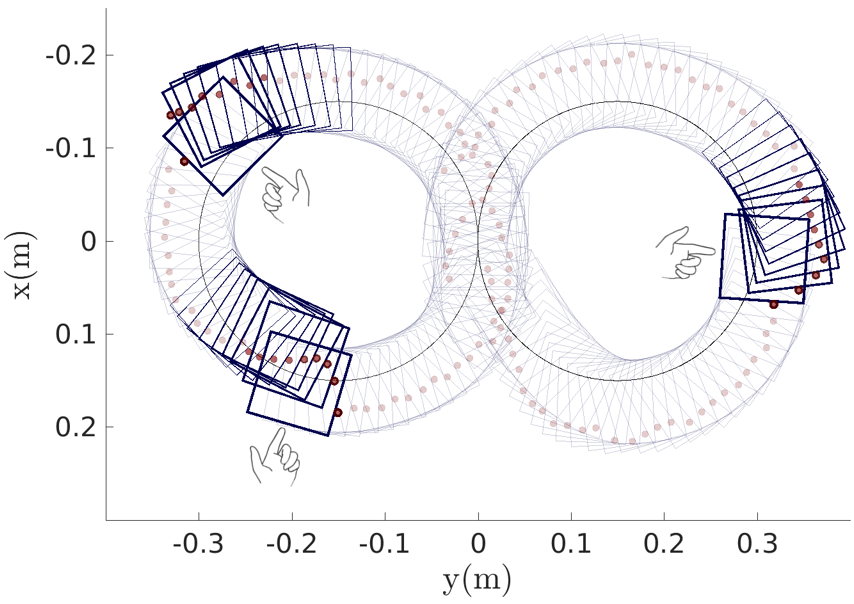}}%
\caption{Point pusher. Closed-loop implementation of the MPC (learned modes) controller, where the goal is to push a square object  about a $8$ shaped trajectory.}
\label{fig:point_pusher_experiments}
\qquad
\subfigure[Tracking of the $8$ track  for  $7$ consecutive laps. The black line represents the desired trajectory and the blue lines track the center of mass of the object.]{%
\label{fig:line_pusher_experiments_first}%
\includegraphics[width=8cm]{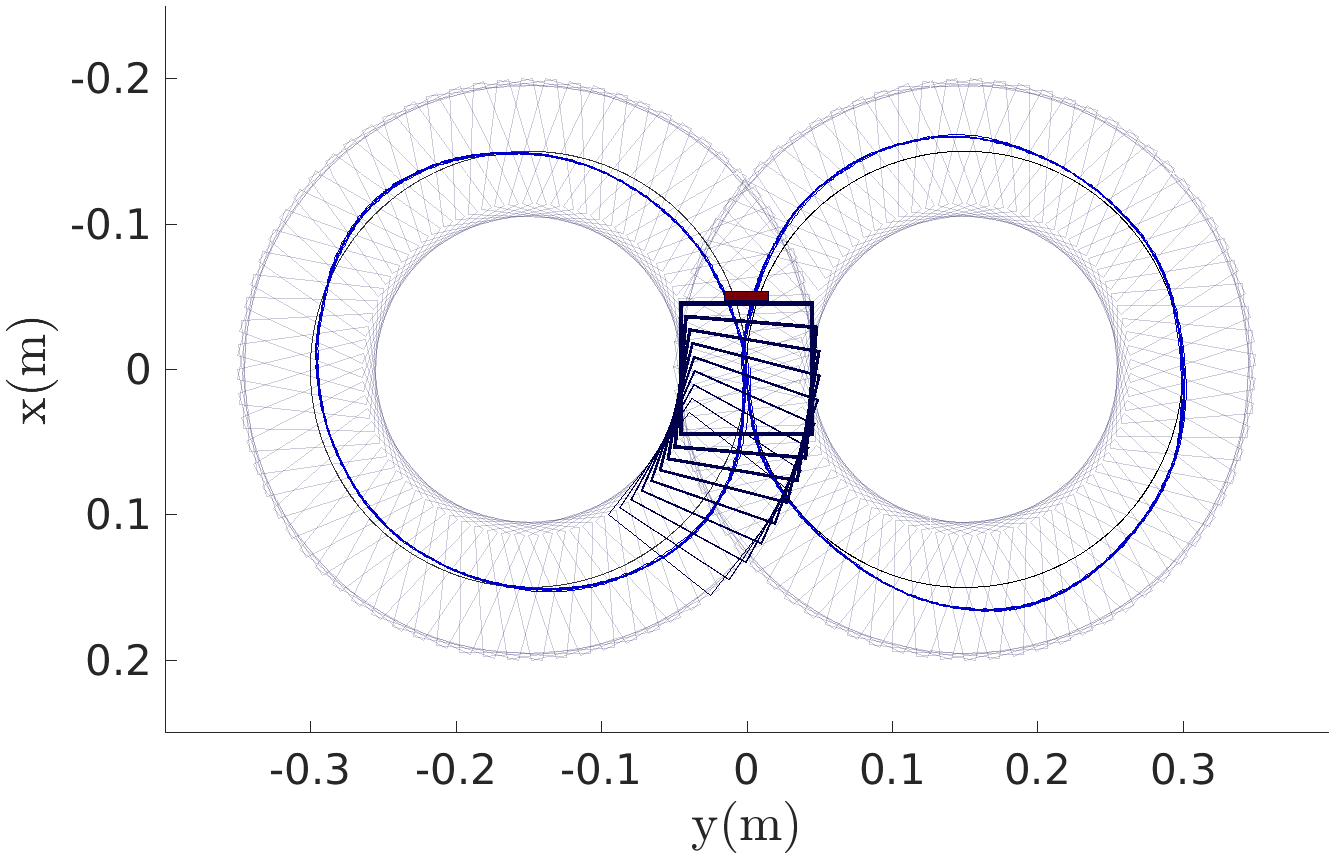}}%
\qquad
\subfigure[Tracking of the $8$ track with  external perturbations for a single laps. The black line represents the desired trajectory and the hand represents the locations and directions in which  the perturbations were applied.]{%
\label{fig:line_pusher_experiments_second}%
 \includegraphics[width=8cm]{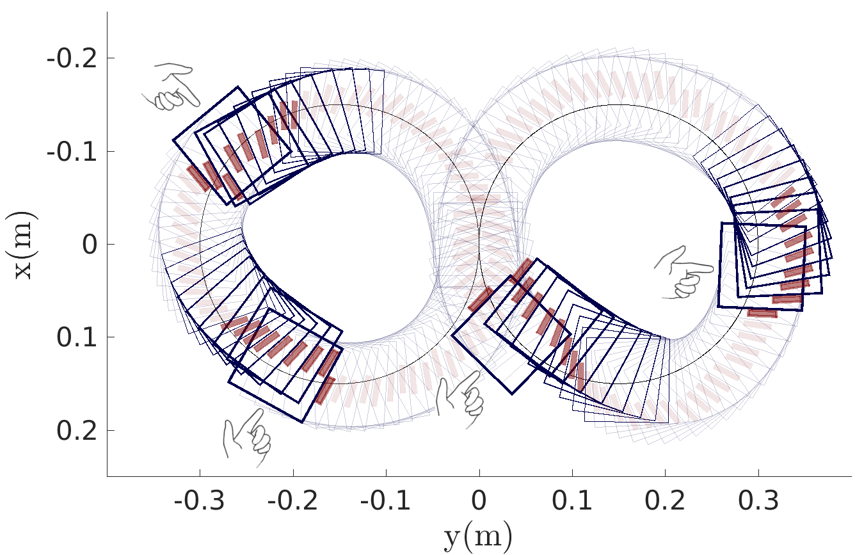}}%
\caption{Line pusher. Closed-loop implementation of the MPC (learned modes) controller, where the goal is to push a square object  about a $8$ shaped trajectory.}
\label{fig:line_pusher_experiments}
\end{figure*}

Figure~\ref{fig:point_pusher_experiments_first}  depicts the robotic point pusher pushing the square object about a $8$ track without any external perturbations for  $7$ consecutive laps. The black line represents the desired trajectory and the blue lines track the center of mass of the object. Although there is a small steady-state error, the controller succeeds in tracking the desired trajectory with accuracy. Figure~\ref{fig:point_pusher_experiments_second}  depicts the robotic point pusher pushing the square object about a $8$ track with external perturbations for  a single lap. The controller quickly succeeds in eliminating the perturbation and returning to the desired trajectory. The novel  convex MPC formulation with learned modes achieves good closed-loop performance and permits a much higher bandwidth ($250$ Hz) than the non-convex MPC (MIQP) formulation ($20$ Hz).

\subsection{Case Study B:  Pushing with Line Contact}
\label{sec:line_pushing}
\begin{figure}[h]
\centering
{
  \includegraphics[width=8.5cm]{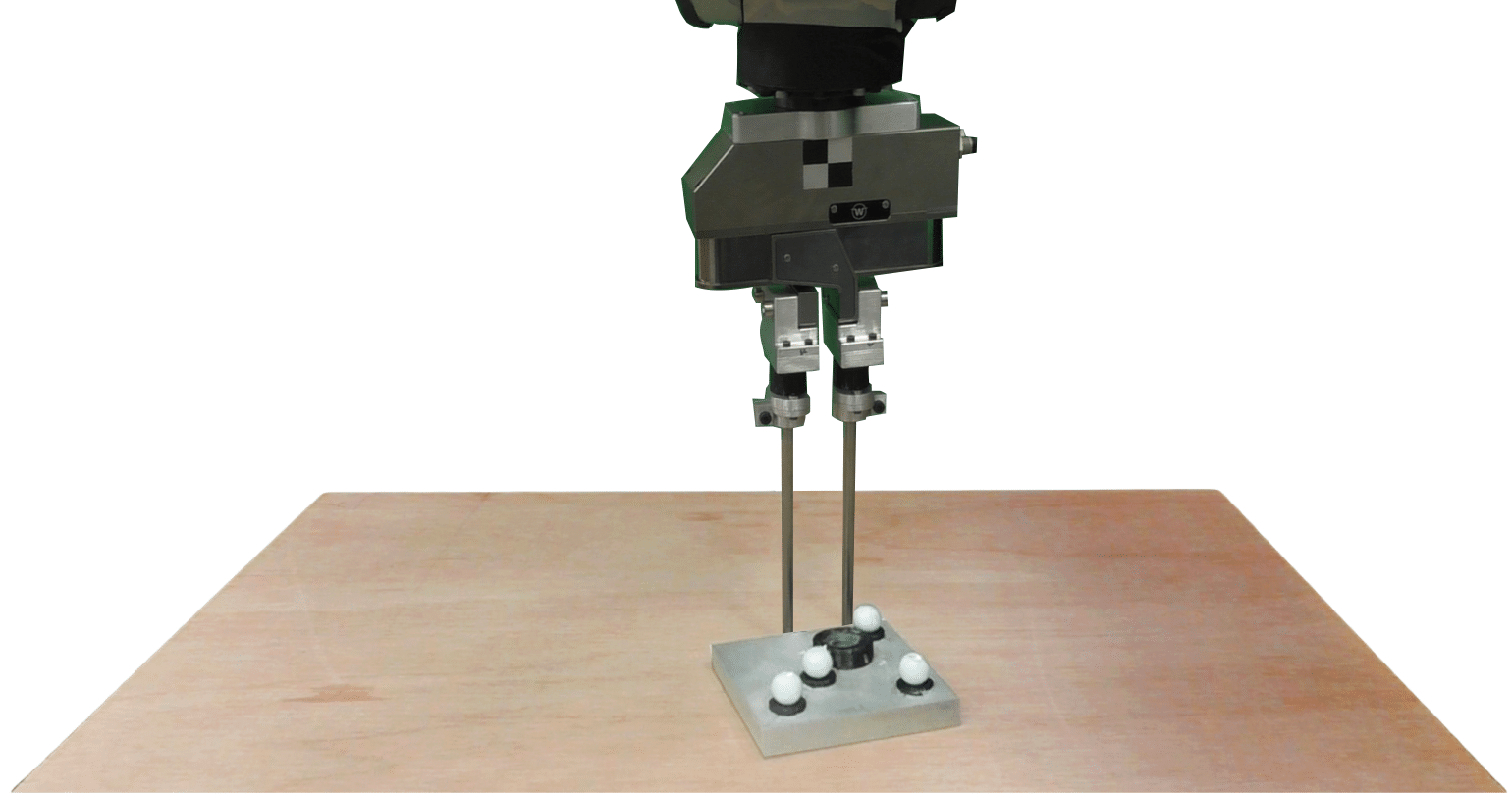} %
}
\centering
\caption{Experimental setup for line pusher.}
\label{fig:line_pusher}
\end{figure}
 
The experimental setup for the planar manipulation task with a line pusher is shown in Fig.~\ref{fig:line_pusher}. The trajectory tracking task is shown in  Fig.~\ref{fig:line_pusher_experiments}, where the goal is to track a 2d trajectory in the shape of an $8$ shaped trajectory defined by two circles of  radii $0.15$ meters at a constant velocity of $v=0.05$ [m/s]. We model the line pusher as $2$ contact points that are constrained to move as a rigid-body, with the position of the center point of pusher denoted as $p_c$ and state vector defined by $\mbf{x} = [x\,\,y\,\,\theta\,\,p_c]^\trans$.  Following a similar approach to that described in Section~\ref{sec:point_pushing}, we train a classifier model using $79410$ labelled data points to predict the optimal mode schedule based on the error state of the system. The  neural network properties used to parametrize the classifier model and the physical properties are related in Table 1 and  2, respectively.  The controller design parameters  are $h=0.3$ seconds, $N=35$, $\mbf{Q} = 10~\text{diag}\{1,1,1,0.1\}$, $\mbf{Q}_N = 2000~\text{diag}\{1,1,1,0.1\}$, and $\mbf{R} =\text{diag}\{1,1,1,1,0.01\}$. We split the prediction horizon  into $8$ parts during which the contact modes are held constant. The number of time steps  associated with each contact mode section $m_m$ is $\{1,5,5,5,5,5,5,4\}$  with the associated weight matrix $\mbf{W} = 0.1~\text{diag}\{0,3,1,1,1,0,0,0\}$.

Figure~\ref{fig:line_pusher_experiments_first}  depicts the robotic line pusher pushing the square object about a $8$ track without any external perturbations for  $7$ consecutive laps. The black line represents the desired trajectory and the blue lines track the center of mass of the object. The steady-state error is less prominent than in the point pusher case, as the line pusher is a more stable system with additional control authority.  Figure~\ref{fig:point_pusher_experiments_second}  depicts the robotic point pusher pushing the square object about a $8$ track with external perturbations for  a single lap. Each time a perturbation is encountered, the pusher reacts to reduce the error by following a fast sliding motion to stabilize the object and then push it back towards the desired trajectory using a sticking phase. The novel MPC (learned modes) controller performs comparably to the MPC (MIQP) formulation and succeeds in tracking the desired trajectory while having significantly faster control bandwidth (200 Hz vs. 15 Hz).

\section{Conclusion}
\label{sec:conclusion}

This paper presents a  methodology for feedback controller design of hybrid dynamical systems. The control formulation is based on a model predictive control approach, where the hybridness and underactuation associated with contact are explicitly enforced as constraints within a  mixed-integer optimization program. 

In order to enable real-time implementation, we  address the combinatorial complexity resulting from the hybrid expansion of contact modes by separating the search for optimal mode schedules (offline) from the search for optimal control inputs (online). This is made possible by formulating the contact mode selection as a supervised learning problem. This approach enables us to train a classifier model offline by harnessing the solutions returned by the mixed-integer optimization program and building a dataset of optimal mode schedules. 

We validate the  controller design methodology on a planar manipulation experimental setup, where it is shown that the proposed convex formulation controller achieves comparable performance as its non-convex alternative, while obtaining a $10$ fold improvement in the control bandwidth. Most importantly, in contrast to mixed-integer MPC control formulations, the online component of the hybrid MPC formulation with learned modes has the potential to extend to more complex dynamical systems with additional contact interactions, as its associated convex program can be efficiently computed in real-time.



\addcontentsline{toc}{section}{References}
{\tiny \bibliographystyle{Include/splncs}}
\bibliography{isrr_bib}

\end{document}